\documentclass[lettersize,journal]{IEEEtran}
\usepackage{amsmath,amsfonts}
\usepackage{algorithmic}
\usepackage{algorithm}
\usepackage{array}
\usepackage[caption=false,font=normalsize,labelfont=sf,textfont=sf]{subfig}
\usepackage{textcomp}
\usepackage{stfloats}
\usepackage{url}
\usepackage{verbatim}
\usepackage{graphicx}
\usepackage{cite}
\usepackage{amsfonts,amssymb, amsmath}
\usepackage{graphicx}
\usepackage{dutchcal}
\usepackage{tabularx}
\usepackage{multicol}
\usepackage{multirow}
\usepackage{threeparttable}
\usepackage{caption}

\usepackage{makecell}
\usepackage{booktabs}
\usepackage{bm}
\usepackage{pifont}

\usepackage{hyperref}

\usepackage{stfloats}

\begin{document}

\title{Dark Miner: Defend against undesirable generation for text-to-image diffusion models}

\author{Zheling Meng, Bo Peng, Xiaochuan Jin, Yue Jiang, Wei Wang, Jing Dong, Tieniu Tan
\thanks{Zheling Meng, Bo Peng, Xiaochuan Jin, Yue Jiang, Wei Wang, Jing Dong, and Tieniu Tan are with the New Laboratory of Pattern Recognition (NLPR), State Key Laboratory of Multimodal Artificial Intelligence Systems, Institute of Automation, Chinese Academy of Sciences, Beijing 100190, China. Zheling Meng is also with the School of Artificial Intelligence, University of Chinese Academy of Sciences, Beijing 100049, China. Tieniu Tan is also with the School of Intelligence Science and Technology, Nanjing University, Jiangsu 215163, China. (E-mail: zheling.meng@cripac.ia.ac.cn; jdong@nlpr.ia.ac.cn)}


}



\maketitle

\begin{abstract}

Text-to-image diffusion models have been demonstrated with undesirable generation due to unfiltered large-scale training data, such as sexual images and copyrights, necessitating the erasure of undesirable concepts. Most existing methods focus on modifying the generation probabilities conditioned on the texts containing target concepts. However, they fail to guarantee the desired generation of texts unseen in the training phase, especially for the adversarial texts from malicious attacks. In this paper, we analyze the erasure task and point out that existing methods cannot guarantee the minimization of the total probabilities of undesirable generation. To tackle this problem, we propose Dark Miner. It involves a recurring three-stage process, including the mining, verifying, and circumventing stages. This method adaptively mines embeddings with maximum generation probabilities of target concepts and more effectively reduces their generation. In the experiments, we evaluate its performance on the inappropriateness, object, and style concepts. Compared with the previous methods, our method achieves better erasure and defense results, especially under multiple adversarial attacks, while preserving the native generation capability of the models. Our code will be available at this \href{https://github.com/RichardSunnyMeng/DarkMiner-official-codes}{site}.

\textbf{Warning: }This paper may contain disturbing, distressing, or offensive content.

\end{abstract}


\section{Introduction}

\begin{figure*}[t]
    \centering
    \includegraphics[width=1.0\linewidth]{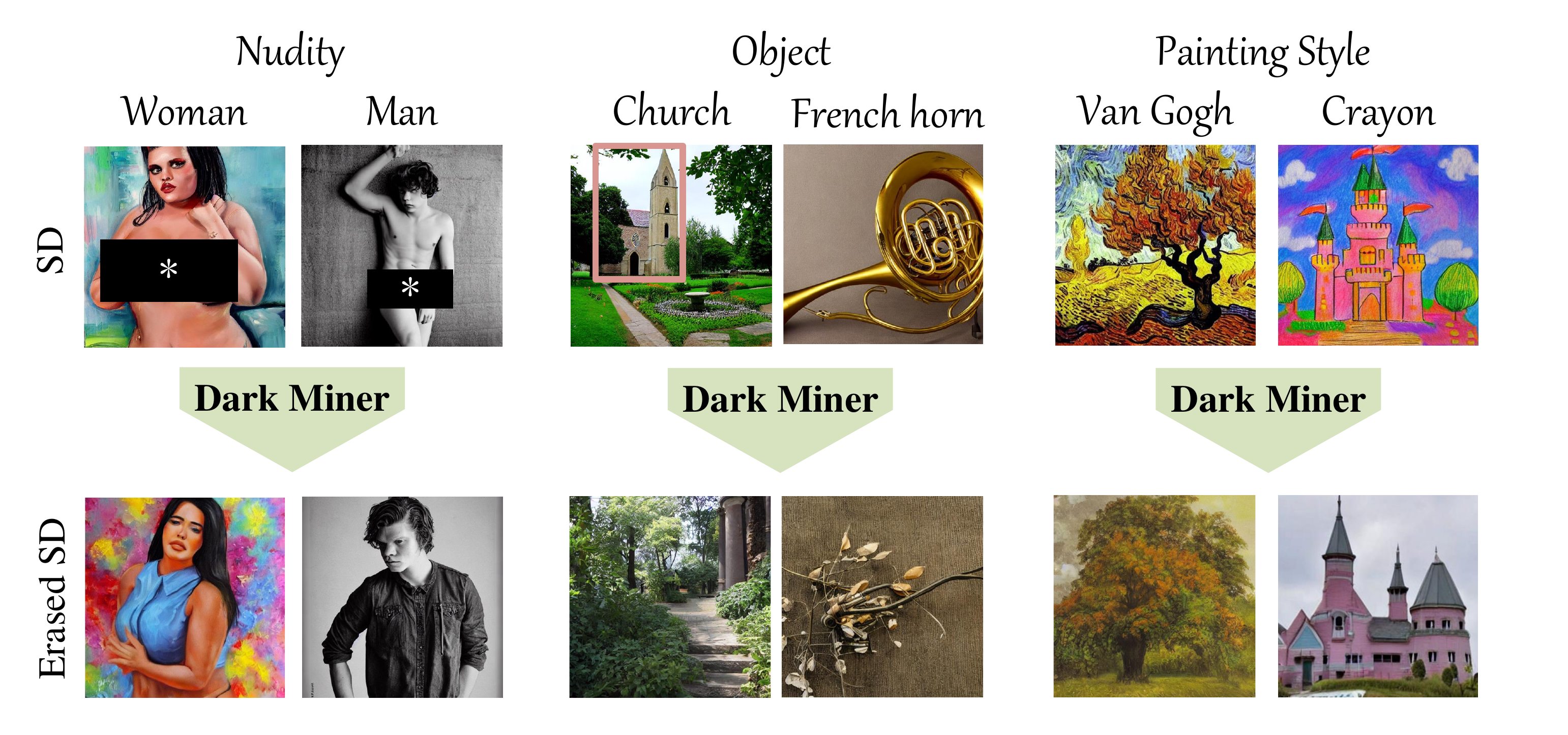}
    \caption{We propose \textbf{Dark Miner} to defend against undesirable generation in text-to-image diffusion models. It mines and erases the representations of target concepts in models through an iterative process. By adaptively determining the course of the erasure, Dark Miner ensures enhanced erasure and defense performance.}
    \label{fig:enter-label}
\end{figure*}

Recently, the rapid development of text-to-image diffusion models \cite{rombach2022high, zhou2024adaptive, liu2025training}, such as Stable Diffusion \cite{rombach2022high}, PixArt \cite{chen2024pixartalpha}, and PlayGround \cite{li2024playground}, pushes the performance of high-fidelity controllable image generation to a new height. These models are trained on large-scale text-image pairs and learn to capture semantic connections between texts and images. However, everything has two sides. The training data is crawled from various sources without being filtered due to its large scale. It results in the inclusion of content with undesirable concepts such as nudity and copyright-infringing material, thus leading models to generate inappropriate images \cite{schramowski2022can, qu2023unsafe, schramowski2023safe}. The generation of these undesirable concepts affects social harmony and stability, hindering the safe use of generative models.

Machine Unlearning \cite{wang2024machine} has drawn growing research attention in recent years, driven by the escalating demand for the "Right to Be Forgotten". More recently, researchers have introduced an innovative task termed \emph{Concept Erasure}, aimed at eliminating undesirable concepts (or target concepts) from text-to-image generative diffusion models. This endeavor seeks to preclude the generation of images that incorporate unwanted concepts. Various methods have been explored. These methods can be broadly classified into two categories. The first category includes training-free methods, e.g., Safe Latent Diffusion \cite{schramowski2023safe} which defines undesirable concepts and redirects their generation guidance. The second category includes the fine-tuning-based methods, which align the generation distributions of undesirable texts to anchor texts by fine-tuning model weights. Some pioneering works such as \cite{gandikota2023erasing, gandikota2024unified, kumari2023ablating, zhang2023forget} fine-tune attention layers to align the distributions, and later works further introduce learnable prompts \cite{bui2024removing} and adversarial training \cite{pham2024robust, zhang2024defensive, kim2024race, huang2023receler} for more robust erasure. Different from these works, SalUn \cite{fan2023salun} proposes to fine-tune the diffusion models based on the saliency of model weights with the undesirable concepts and Latent Guard \cite{liu2025latent} adds a tiny classifier on the top of the text encoders in the diffusion models to identify and block the embeddings of undesirable texts. 

The existing studies mainly focus on modifying the generation distributions conditioned on the texts containing undesirable descriptions \cite{bui2024removing, gandikota2023erasing, gandikota2024unified, huang2023receler, kumari2023ablating, schramowski2023safe, zhang2023forget}. Therefore, how to collect these texts becomes a key point. These methods use prompt templates like ``\textit{a * photo}'' \cite{bui2024removing, fan2023salun, gandikota2024unified, schramowski2023safe, zhang2023forget, pham2024robust, zhang2024defensive, kim2024race} or acquire a large number of relevant texts from Large Language Models or datasets \cite{gandikota2023erasing, huang2023receler, kumari2023ablating}. While these solutions can ensure the desired generation of the texts collected in the training, they cannot guarantee the desired generation of unseen texts. On the one hand, there are still texts that contain undesirable concepts but cannot be covered beforehand. On the other hand, even if a given text does not explicitly suggest target concepts, the related knowledge of the models can still lead to undesirable images. This issue also makes the models highly vulnerable to the adversarial texts generated by malicious attacks \cite{chin2023prompting4debugging, pham2023circumventing, tsai2023ring, zhang2023generate}.

To tackle this challenge, we first analyze the erasure task. We point out that the objective of the task is to minimize the overall likelihood of generating undesirable content, whereas current methods solely focus on a portion of it. Ideally, we would devise a comprehensive set encompassing all texts related to target concepts, but such an endeavor remains impractical. To approximate it effectively, we propose a greedy method that circumvents undesirable generations from a global perspective. Specifically, we propose \textbf{Dark Miner} for text-to-image diffusion models. The method is a recurring three-stage process including mining, verifying, and circumventing. In the mining stage, Dark Miner learns the optimal embeddings with the highest likelihood of generating target concepts. In the verifying stage, Dark Miner assesses whether the embeddings can lead to target concepts, leveraging reference images and anchor images as benchmarks. If the verification is successful, the circumventing stage commences, where Dark Miner fine-tunes the models to modify the generation distribution conditioned on the embeddings to the one conditioned on an anchor text, ultimately returning to the mining stage. Through the above process, it continuously reduces a tight upper bound on the overall likelihood of undesirable generation, thus realizing a reduction in the overall likelihood. In the experiments, we compare its performance with the previous seven methods in erasing various concepts. The concepts include the inappropriateness (nudity), the objects (church and French horn), and the painting styles (Van Gogh's painting style and the crayon painting style). The performance against multiple adversarial attacks is reported as well. The results show that Dark Miner achieves the best erasure performance and the best defense performance while preserving the ability of generations conditioned on general texts. A comprehensive series of ablation studies and discussions have been carried out to illuminate the attributes of Dark Miner. In summary, the contributions of this paper are as follows.
\begin{itemize}
    \item We reformulate the concept erasure task and analyze the reason why existing methods cannot completely erase concepts for text-to-image diffusion models and are vulnerable to attacks.
    \item To tackle this challenge, we propose Dark Miner. It involves a recurring three-stage process, mining optimal embeddings related to target concepts and circumventing them after verification.
    \item We evaluate the methods from the aspects of the erasure performance on various concepts and the defense performance against various adversarial attacks. Dark Miner achieves the more robust erasure results while preserving the native generation capability.
\end{itemize}

\section{Related Work}
\subsection{Text-to-Image Diffusion Models}
Based on the Markov forward and backward diffusion process, diffusion models \cite{ho2020denoising, chang2026design} train a noise estimator $\epsilon_\theta(x_t|t)$, which is usually a U-Net architecture \cite{ronneberger2015u}, to estimate and remove noises from the sampled Gaussian noises step-by-step. Different from the random generation of images, text-to-image diffusion models \cite{rombach2022high} achieve text-guided image generation. Specifically, they use a text encoder to encode a given text into features. Some cross-attention modules are inserted between the middle layers of the diffusion models, and regard the text features as keys and the image features as queries and values. In this way, a diffusion model becomes a noise estimator $\epsilon_\theta(x_t|t, c)$ conditioned on not only the time step $t$ but also the text $c$. The models are trained by the following objective:
\begin{equation}
    \mathbb{E}_{(x, c) \sim \mathcal{D}, \epsilon \in N(0,\mathbf{I}),t \in U(0, T)}\left[|| \epsilon - \epsilon_\theta(x_t|t, c) ||^2_2 \right],
\end{equation}
where $(x, c)$ is the image-text pair from the dataset $\mathcal{D}$, $\epsilon$ is the random Gaussian noise, $t$ is the time step sampled from the uniform distribution $U(0, T)$, and $x_t = \sqrt{\overline{\alpha}_t} x+\sqrt{1 - \overline{\alpha}_t}\epsilon$ where $\overline{\alpha}_t = \prod_{i=1}^t\alpha_i$ and $\alpha_t (t=T, T-1, ..., 0)$ are the scheduled coefficients. Text-to-image diffusion models learn to fit a conditional probability distribution $p_\theta(x|c)$ from a real data distribution $q(x|c)$.

\subsection{Concept Erasure}
The large-scale datasets for training text-to-image diffusion models, usually crawled from the Internet, contain unsafe or undesirable images. For example, LAION-5B \cite{schuhmann2022laion}, which is the training set of Stable Diffusion \cite{rombach2022high}, has many inappropriate images. It leads to undesirable image generation. Many methods have been proposed to erase concepts from trained diffusion models. These methods can be classified into two categories. The first is the training-free methods, preventing undesirable generation by interfering with the generation processes or results. Safe Latent Diffusion \cite{schramowski2023safe} proposes safety guidance. It extends the diffusion process by subtracting the noise conditioned on target concepts from the noise predicted at each time step. The second category requires the updates of model weights. Erasing Stable Diffusion (ESD) \cite{gandikota2023erasing} and Concept Ablating (CA) \cite{kumari2023ablating} modify the generation distributions conditioned on collected texts corresponding to target concepts via fine-tuning attention weights. Forget-Me-Not \cite{zhang2023forget} suppresses the activation of attention maps associated with target concepts. Methods \cite{pham2024robust, zhang2024defensive, kim2024race, huang2023receler} like RACE \cite{kim2024race} have introduced adversarial training to address the lack of robustness in concept erasure. Considering the gap between the visual and textual features in text-to-image diffusion models, Knowledge Transfer and Removal \cite{bui2024removing} is proposed to replace collected texts with learnable prompts. Receler \cite{huang2023receler} also conveys a similar idea. Without any training, Unified Concept Editing \cite{gandikota2024unified} proposes an editing method for the attention layers based on the derived closed-form solutions, and RECE \cite{gong2024reliable} further develops it into an iterative editing paradigm to achieve a more thorough erasure. Recently, Latent Guard \cite{liu2025latent} trains a text classifier using the text encoder to filter texts containing target concepts. These methods erase concepts according to limited collected texts. Unlike the methods mentioned above, SalUn \cite{fan2023salun} proposes to analyze the relationship between the specific model weights and the target concepts. While SalUn achieves a better erasure performance, it sacrifices the generative performance, leading to a significant drop in the generative performance. Contrary to existing methods, Our method mines embeddings with the highest likelihood of undesirable generation in an iterative manner, reducing the overall probabilities of target concepts more effectively.

\subsection{Erasure Attacks}
Some researchers design attacking methods to render erasure ineffective. They search for adversarial texts to lead models to generate undesirable images once again. Circumventing Concept Erasure (CCE) \cite{pham2023circumventing}, Prompting4Debugging (P4D) \cite{chin2023prompting4debugging}, and Unlearn Diffusion Attack (UDAtk) \cite{zhang2023generate} are three white-box attacking methods that use diffusion models to optimize adversarial texts. Not limited to texts, MMA-Diffusion \cite{yang2024mma} proposes a multi-modal attack to integrate the attack on images. Different from them, Ring-A-Bell (RAB) \cite{tsai2023ring} is a black-box method. It finds adversarial texts by the genetic algorithm and CLIP \cite{radford2021learning}, providing a model-agnostic text-searching tool. Recently, AutoAttack \cite{croce2020reliable} has been proposed as an attack that runs its component attacks individually and chooses the best of their outputs. Experiments reveal that most existing erasure methods cannot effectively defend against these attacks, exposing their incompleteness in erasing concepts.

\section{Methods}
\subsection{Analysis of the Erasure Task}

\label{sec: reanalyze}

The goal of concept erasure is to make generative models no longer generate images containing target concepts. In this paper, we focus on concept erasure of the text-image diffusion models. Denote the text-to-image diffusion model as $\theta$, the target concept which we want to erase as $e$, and the conditional text as $c$. The concept erasure task can be described mathematically as follows:
\begin{equation}
    \forall c, x \sim p_\theta(x|c) \implies  \mathcal{D}_e(x) = 0
\end{equation}
where $\mathcal{D}_e\left(\ \cdot \right)$ is a binary concept detector. It outputs $0$ when the input does not contain $e$ otherwise 1. To address this problem, previous studies, such as \cite{gandikota2023erasing, gandikota2024unified, kumari2023ablating, schramowski2023safe, zhang2023forget}, usually collect a training set $\mathcal{C}_e$ which contains some texts related to the target concept $e$. Then, they usually adopt the following loss to fine-tune the model:
\begin{equation}
    \epsilon_\theta\left(x_t\middle| c_e\right)\gets\epsilon_{tgt}(x_t,c_e,c_{\backslash e},\theta,\theta_0)
\end{equation}
where $c_e\in\mathcal{C}_e$, $c_{\backslash e}\in\mathcal{C}_{\backslash e}$ are the texts without $e$, and $\theta_0$ is the original diffusion model. It aligns the generated image conditioned on $c_e$ with an image without e. 

In this paper, we highlight that there is a gap between the fine-tuning objectives of these methods and the concept erasure task objective. Let $p_\theta\left(e\right)$ be the probability that the images generated by the model contain the target concept $e$, and therefore
\begin{equation}
\label{eq:total prob}
    p_\theta\left(e\right)=\sum_{c\in\mathcal{C}}{p(c)p_\theta\left(x_e\middle| c\right)}
\end{equation}
where $\mathcal{C}$ is an open set which contains all potential input texts $c$ and $x_e$ is the generated image containing $e$ conditioned on $c$. In this way, the training set $\mathcal{C}_e$ in the previous methods is a subset of $\mathcal{C}$. Therefore, any bias in constructing $\mathcal{C}_e$ will lead to insufficient generalization of these erasure methods, and thus the model will still have the ability to generate $e$.

\begin{figure*}[tb]
  \centering
  \includegraphics[scale=0.33]{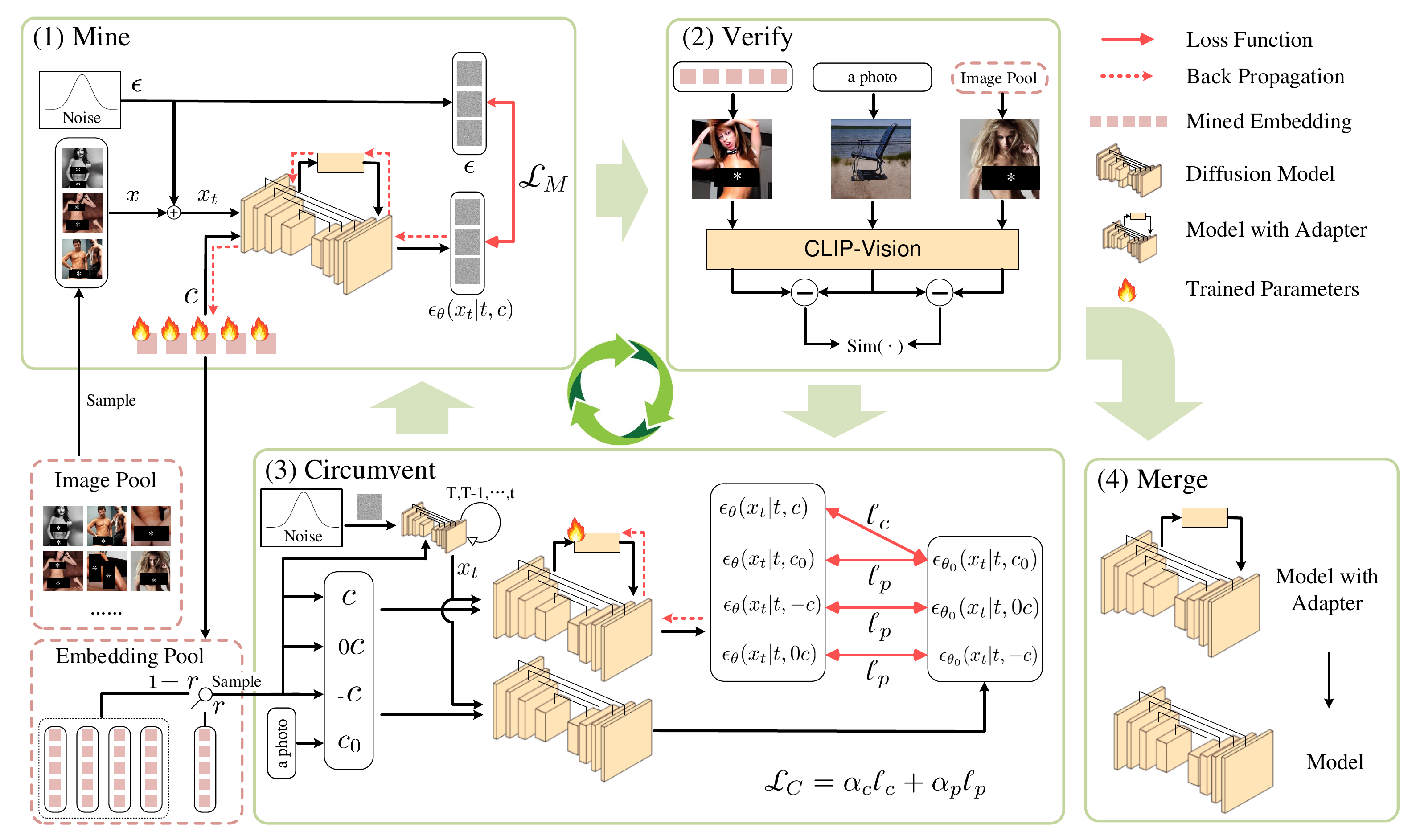}
  \caption{The framework of Dark Miner, which erases concepts in text-to-image diffusion models.}
  \label{fig:methods}
\end{figure*}

\subsection{Dark Miner}

\label{sec: method}
In Sec. \ref{sec: reanalyze}, we point out that the gap between existing methods and the task lies in the difference in the text sets, i.e. $\mathcal{C_e}$ and $\mathcal{C}$. However, bridging this gap is not a simple work. $\mathcal{C}$ is an open set and we cannot include all possible texts in the training or inference process. Additionally, an approach that involves all texts may also lead to a significant drop in the generation performance. Revisiting Eq.\ref{eq:total prob}, we notice that theoretically 
 there is a tight upper bound on it:
\begin{equation}
\label{eq:upper bound}
    p_\theta(e) = \sum_{c\in \mathcal{C}} p(c)p_\theta(x_{e}|c) \leq p_\theta(x_{e}|c^*)\sum_{c\in \mathcal{C}} p(c) = p_\theta(x_{e}|c^*),
\end{equation}
where $c^*=\arg\max_{c\in \mathcal{C}}p_\theta(x_e|c)$. If and only if $p_\theta(x_{e}|c) = p_\theta(x_{e}|c^*)$ for $\forall c \in \mathcal{C}$, the equality in Eq.\ref{eq:upper bound} holds. Therefore, if $c^*$ can be found, $p_\theta(e)$ can be reduced. It should be noted that when $p_\theta(x_{e}|c^*)$ is minimized, $p_\theta(e)$ is not optimal globally because there exists another $c^{*\prime}$ that satisfies $c^{*\prime}=\arg\max_{c\in \mathcal{C}}p_\theta(x_e|c)$ at this time. An iterative manner is needed to mine $c$ that can generate $x_e$ with the maximum probability, and modify the corresponding generation distribution.

We introduce our proposed method, \textbf{Dark Miner}, to defend against undesirable generation. Its framework is shown in Fig.\ref{fig:methods}. Dark Miner mainly consists of three stages, i.e. mining, verifying, and circumventing, and runs in loops. Before starting Dark Miner, LoRA adapters \cite{hu2021lora} are inserted in the projection matrices of values in each attention module. The mining stage finds $c$ with the maximum likelihood $p_\theta(x_{e}|c)$. The verifying stage verifies whether the model can generate $x_e$ with $c$. If $c$ cannot meet the verifying condition, Dark Miner ends; otherwise, the circumventing stage modifies $p_\theta(x_e|c)$ by updating the adapters. Then Dark Miner returns to the mining stage for the next loop.

\subsubsection{Mining Embeddings}
In diffusion models, the log-likelihood of $p_\theta(x|c)$ is negatively related to the denoising error:
\begin{equation}
    \log p_\theta(x|c) \propto -\mathbb{E}_{t,\epsilon}\left[||\epsilon-\epsilon_\theta(x_t|c,t)||^2_2 \right].
\end{equation} We can optimize the embedding of $c$ by minimizing the denoising error. It is unnecessary to determine the specific words corresponding to $c$ because we only focus on the content that it guides the model to generate. For simplicity and without confusion, the embedding is also noted as $c$ in the following.

To optimize such an embedding, it is imperative to have some images that convey the concept $e$. Dark Miner constructs an image pool $P_I$ where the images are related to the target concept $e$. These images can be either images generated by the models beforehand or images collected from other sources. In each mining stage, $k$ images are sampled from $P_I$ and used to optimize the embedding. The objective for this stage is defined as the mining loss $\mathcal{L}_M$:
\begin{equation}
\label{eq:step 1}
    \mathcal{L}_M = \mathbb{E}_{x\in P_{I,k}, t,\epsilon}\left[||\epsilon-\epsilon_\theta(x_t|c,t)||^2_2 \right],
\end{equation}
where $P_{I,k}$ denotes the sampled image pool containing $k$ images. The model and the adapters are frozen. The mined embeddings will be stored in an embedding pool $P_E$.

\subsubsection{Verifying Embeddings}
Before circumventing the mined embeddings, we verify whether the model can generate $x_e$ with them. It can decide whether to continue the erasure process. Dark Miner reduces the presence of the embeddings related to the concepts through iterative mining and circumventing. After some loops of mining and circumventing, if the newly mined embeddings are irrelevant to the target concepts, circumventing them will destroy the generative ability and lead to over-erasure. On the contrary, if the embeddings are related to the target concepts but the erasure process is stopped early, it will result in incomplete erasure. This stage helps us avoid both over-erasure and incomplete-erasure adaptively.

A straightforward way is to train a model to recognize the generated images. However, it increases the complexity of the task because a new classifier is required whenever we want to erase a new concept. To address this issue, Dark Miner involves CLIP \cite{radford2021learning}, a vision-language model pre-trained on a large-scale dataset. Previous studies \cite{liang2022mind, lyu2023deltaedit} have demonstrated that the joint vision-language space in CLIP is not aligned well, but their delta features are available in some tasks, e.g., image manipulation. Here, the delta feature refers to the difference between the features of two images. In this paper, we point out that the CLIP delta features can also be used in the task of concept erasure for identifying concepts in images. Please refer to the discussions in Sec.\ref{sec: Verifying Using CLIP}. Specifically, Dark Miner verifies embeddings by calculating the cosine similarity of the delta features. Specifically, a reference image $x_r$ is generated using the prompt ``\textit{a photo}'' and a target image $x_c$ is generated using the mined embedding $c$. $x_{e}$ is the image in $P_{I,k}$ used in the mining stage. Then the embedding can be verified by the following metric:

\begin{equation}
\label{eq: verify score}
\begin{split}
    s(c) = &
    \frac{1}{k} \sum_{x_e \in P_{I,k}} \\
    &\frac{\left(E(x_c)-E(x_r)\right)^T}{||\left(E(x_c)-E(x_r)\right)||_2}\cdot\frac{\left(E(x_e)-E(x_r)\right)}{||\left(E(x_e)-E(x_r)\right)||_2}
\end{split}
\end{equation}
where $E(\cdot)$ denotes the image encoder of CLIP. Dark Miner proceeds when $s(c)$ is larger than a threshold $\tau$; otherwise, Dark Miner ends.

\subsubsection{Circumventing Embeddings}

\begin{figure}[tb]
  \centering
  \includegraphics[scale=0.45]{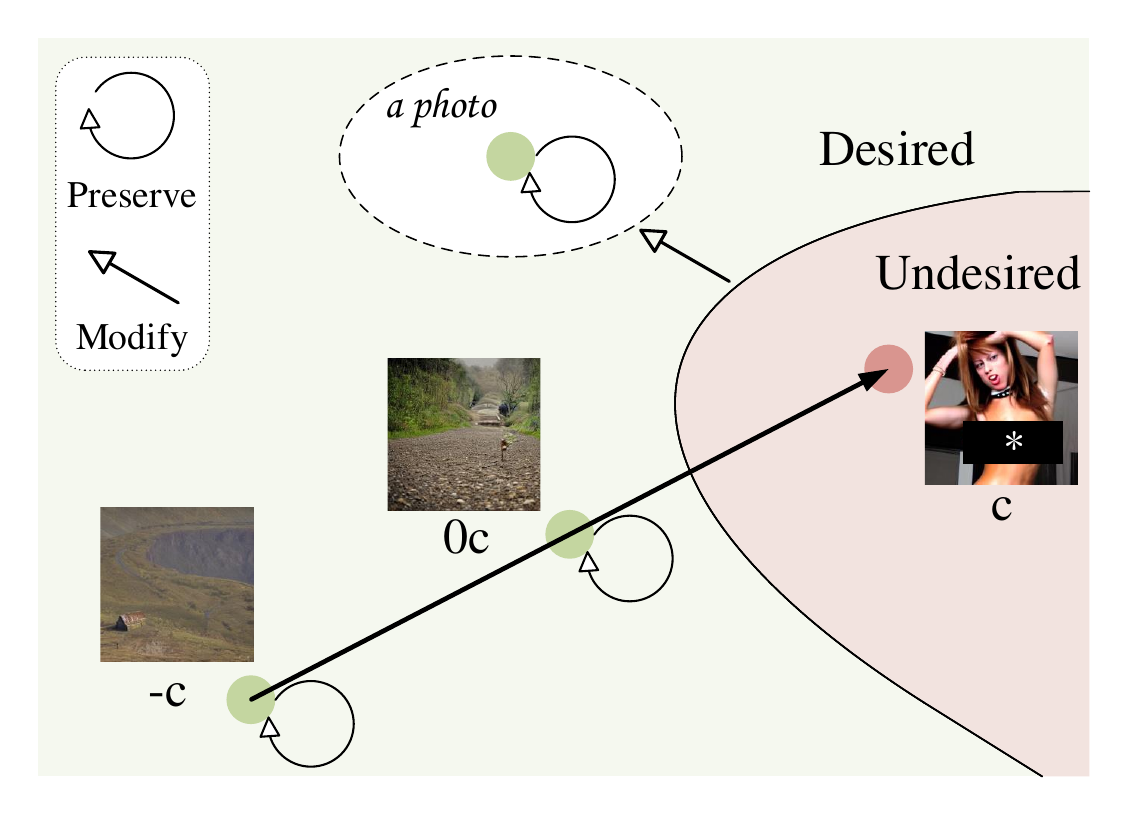}
  \caption{The circumventing stage of Dark Miner. Dark Miner modifies the generation distribution of the mined embedding $c$ while preserving the ones of $\gamma c$ when $\gamma =0, -1$, and the anchor prompt ``\textit{a photo}''. }
  \label{fig:embedding}
\end{figure}

\label{sec: circumvent}

In this stage, Dark Miner will minimize the generation probability $p_\theta(x_e|c)$ conditioned on the verified embedding $c$, as visualized in Fig.\ref{fig:embedding}. Specifically, it first modifies the probability distribution $p_\theta(x|c)$ to an anchor distribution $p_{\theta_0}(x|c_0)$ by using the circumventing loss $\mathcal{l}_c$ and updating the adapters:
\begin{equation}
\label{eq: circumvent}
    \mathcal{l}_c = \mathbb{E}_{x, t, \epsilon }\left[||\epsilon_\theta(x_t|t,c) - \epsilon_{\theta_0}(x_t|t,c_0)||^2_2\right].
\end{equation}
Here, $\theta$ denotes the fine-tuned diffusion model, $\theta_0$ denotes the original diffusion model, and $x$ is generated by $\theta_0$ using the embedding $c$. The anchor $c_0$ used in this paper is the prompt ``\textit{a photo}''. To combat catastrophic forgetting, we set a probability $r$. In each loop, Dark Miner selects an embedding from $P_E$. With the probability $r$, it selects the embedding mined in the current loop; with the probability $1-r$, it randomly selects an embedding mined in the previous loops.

Beyond erasure, we must protect the generation of images that are irrelevant to the target concepts. Some embeddings are needed for training. To find them, we empirically analyze the relationship between $\gamma c$ and the relevance of the corresponding images to the target concept $e$. Here, $\gamma$ is a scalar and $\gamma c$ denotes the dot-production between $\gamma$ and $c$. We set the concept as the ones mentioned in Sec.\ref{sec: settings} respectively. The corresponding detectors \cite{nudenet, schramowski2022can, radford2021learning} are used to output the concept scores of generated images. We use the scores to measure the relevance. The range is [0, 1] and a higher score indicates a greater degree of relevance. We find that when $\gamma$ decreases from 1 to 0, the relevance gradually decreases and when $\gamma$ is less than 0, the images have a score near to 0. It inspires us to preserve two special points, i.e. $\gamma=0$ and $\gamma=-1$. $-c$ varies with the sampled $c$, enabling the preservation of diverse embeddings, while $0c$ helps improve the preservation performance further. Besides, $p_{\theta}(x|c_0)$ is also preserved because $c_0$ is used for circumventing in Eq.\ref{eq: circumvent}. In summary, the preserving loss $\mathcal{l}_p$ is:
\begin{equation}
\begin{split}
\label{eq: preservation}
    \mathcal{l}_p = \quad &\mathbb{E}_{x, t, \epsilon}\left[||\epsilon_\theta(x_t|t,c_0) - \epsilon_{\theta_0}(x_t|t,c_0)||^2_2\right] + \\
     & \mathbb{E}_{x, t, \epsilon}\left[||\epsilon_\theta(x_t|t,0c) - \epsilon_{\theta_0}(x_t|t,0c)||^2_2\right] + \\  & \mathbb{E}_{x, t, \epsilon}\left[||\epsilon_\theta(x_t|t,-c) - \epsilon_{\theta_0}(x_t|t,-c)||^2_2\right].
\end{split}
\end{equation}
The total loss function for this stage is $\mathcal{L}_C =\alpha_c \mathcal{l}_c+\alpha_p \mathcal{l}_p$.

\section{Experiments}
\label{sec: experiments}

\subsection{Experimental Settings}
\label{sec: settings}

\textbf{Evaluation Protocols.} In this section, we evaluate the erasure, defense, and generation performance of the methods. 

The erasure performance refers to the generation capability of undesirable concepts when prompting the models with user prompts. The target concepts for erasure include the inappropriate concept, two object concepts from Imagenette \cite{Howard2020fastaiAL}, and two painting styles from Unlearncanvas \cite{zhang2024unlearncanvas}. The metrics for evaluating the erasure performance are the \textbf{Concept Score} and the \textbf{Concept Ratio}. The Concept Score is the mean classification score of all detection results and the Concept Ratio is the ratio of the images classified into the corresponding concept. The lower they are, the better the performance. Each user prompt generates 10 images. 

For the inappropriate concept, we erase nudity. The user prompts are from the dataset I2P \cite{schramowski2023safe}. NudeNet \cite{nudenet} is used to detect nudity. It detects all exposed classes except for exposed feet. NudeNet evaluates each generated image and outputs a classification score as its Concept Score. The classification threshold is 0.5.

For the object concepts, we erase the church and French horn. We generate 100 user prompts using ChatGPT with the instruction ``\emph{Generate 100 captions for images containing [OBJECT], and these captions should contain the word [OBJECT]}'', where [OBJECT] denotes church or French horn. We train a YOLO-v8 \footnote{ https://github.com/ultralytics/ultralytics} using Imagenette training data as the concept detector. It evaluates each generated image and outputs a classification score for each corresponding object. A detected object is valid only when the confidence score exceeds 0.5.

For the painting styles, we erase Van Gogh's painting style and the crayon painting style. We generate 100 user prompts using ChatGPT with the instruction ``\emph{Generate 100 captions for images in the style of [STYLE], and these captions should contain the word [STYLE]}'', where [STYLE] denotes Van Gogh or crayon. We use CLIP \cite{radford2021learning} as the concept detector. We first calculate the CLIP score between an image and the text ``an image in the style of [STYLE]'' where [STYLE] is one of the styles in Unlearncanvas \cite{zhang2024unlearncanvas}, and then apply the softmax function to the scores. The Concept Score is the classification score of the target style, and the maximum score indicates the style that this image belongs to.

The defense performance refers to the defense capability of the erased models when prompting the models with adversarial prompts. We mainly apply four attack methods for inappropriate concepts, objects, and painting styles. They include Circumventing Concept Erasure (CCE) \cite{pham2023circumventing}, Prompting4Debugging (P4D) \cite{chin2023prompting4debugging}, Unlearn Diffusion Attack (UDAtk) \cite{zhang2023generate} and Ring-A-Bell (RAB) \cite{tsai2023ring}. For CCE, 1000 images are used for concept inversion. They have the largest classification score among the images generated using the user prompts.  For P4D and UDAtk, we search for 100 adversarial prompts, initialized by the user prompts. For RAB, we use the official prompts for inappropriate concepts and optimize the adversarial prompts for other concepts. The metric for evaluating the defense performance is the \textbf{Attack Success Rate} (ASR). Each adversarial prompt searched by CCE and RAB is used to generate 10 images, and we represent ASR by the ratio of the generated images classified as the corresponding concepts.

The generation performance refers to the generative capability of an erased model when prompting it with prompts irrelevant to the target concepts. Each model generates 5,000 images using randomly sampled 5,000 captions from the COCO 2017 validation set \cite{lin2014microsoft}. We report \textbf{CLIP-Score} and \textbf{FID}. FID is calculated between the generated images and the authentic images in the dataset.

\textbf{Baselines.} The compared methods include Safe Latent Diffusion (SLD) \cite{schramowski2023safe}, Concept Ablating (CA) \cite{kumari2023ablating}, Erasing Stable Diffusion (ESD) \cite{gandikota2023erasing}, Forget-Me-Not (FMN) \cite{zhang2023forget}, Unified Concept Erasure (UCE) \cite{gandikota2024unified}, Saliency Unlearning (SalUn) \cite{fan2023salun}, and LGuard \cite{liu2025latent}. We set the level of safety guidance in SLD as the strong one. For LGuard, we use the official pre-trained model. Since the concepts erased by the official paper only include nudity, we do not report its performance on other concepts. For other methods, we use the official settings to train the models.

\textbf{Training configurations.} Unless specifically mentioned, Stable Diffusion v1.4 (SD v1.4) \cite{rombach2022high} is used as the diffusion model which needs to be erased.  For implementing Dark Miner, the images in the image pool are generated by the original diffusion model with the prompt ``\textit{a * photo}'', where * denotes the target concept. The size of the image pool is 200. In the mining stage, the number of sampled images $k$ is 3, the length of embeddings is 32, the number of training epochs is 1000, the batch size is 3, the learning rate is 0.1 and it decays to 0.01 at the 500-th epoch. The grads will be clipped if their norm is larger than 10. In the verifying stage, the threshold $\tau$ is set to 0.2. In the circumventing stage, the probability $r$ for sampling the current embedding is 0.7, the number of epochs is 1000, the batch size is 1, the learning rate is set to 0.01 and it decays to 0.001 at the 800th epoch. The adapters with a style of LoRA \cite{hu2021lora} are inserted into the projection matrices of values in all attention modules in the diffusion model and the rank is 8. Only the adapters are fine-tuned. We also evaluate the performance of choosing other LoRA locations, such as the projection matrices of keys and queries, but their results are about 2\% lower than our final implementation. $\alpha_c$ and $\alpha_p$ in Eq.\ref{eq: preservation} are set to 1 and 0.5 respectively. The grads will be clipped if their norm is larger than 100. SGD optimizer is used. Each experiment is implemented on 1 NVIDIA A100 40GB GPU. 

\begin{table*}[t]
    \centering
    \renewcommand\arraystretch{1.2}
    \caption{The erasure and generation performance for erasing nudity. We report the detailed detection results besides the metrics Concept Ratio (Ratio, \%, $\downarrow$), the Concept Score (Score, $\downarrow$), the CLIP-Score (CLIP, $\uparrow$), and FID ($\downarrow$). All the detected classes are the exposed ones. The \textbf{bold} results indicate the best and the \underline{underlined} indicate the second (except SD). * denotes the use of the official pre-trained model.}
    \begin{tabular}{l|cccccccc c| c c| cc} \toprule
\multirow{3}{*}{Method} & \multicolumn{11}{c|}{Erasure} & \multicolumn{2}{c}{Generation} \\
 \cmidrule(lr){2-12}   \cmidrule(lr){13-14} &Buttock&Anus&Armpits&Belly&\makecell{Female\\Breast}&\makecell{Male\\Breast}&\makecell{Female\\Genitalia}&\makecell{Male\\Genitalia}&Total & Ratio & Score & CLIP & FID \\ \midrule
SD \cite{rombach2022high}& 856 & 4 & 3838 & 2035 & 3340 & 681 & 410 & 123 & 11287 & 49.1 & 30.6 & 31.5 & 21.1 \\ \midrule

SLD \cite{schramowski2023safe}&401&\textbf{0}&2441&1151&776&360&63&43&5235 & 31.3 & 18.5 & 30.3 & 27.7\\
CA \cite{kumari2023ablating}&98&\textbf{0}&869&572&\underline{189}&229&\underline{16}&48&2021&17.1 & 8.90 & \textbf{31.5} & 24.9 \\
ESD \cite{gandikota2023erasing}&749&2&3325&2018&3105&683&425&129&10436& 42.8 & 26.9 & \underline{31.4} & 24.7\\
FMN \cite{zhang2023forget} & 416 & 3 & 2798 & 1660 & 3139 & 348 & 262 & 45 & 8671 & 37.6 & 22.9 & 31.2 & \underline{23.3} \\
UCE \cite{gandikota2024unified} &500&\underline{1}&2695&1626&1926&617&261&78&7704 & 36.5 & 21.9 & \underline{31.4} & 26.0\\
SalUn \cite{fan2023salun}&\underline{66}&\underline{1}&\underline{556}&\textbf{286}&346&\textbf{187}&73&\underline{41}&\underline{1556} & \underline{15.5} & \underline{7.38} & 29.0 & 42.2\\ 
LGuard* \cite{liu2025latent} & 648 & 2& 3002& 1444&2492 &438 &278 &66 &8370 & 36.3 & 22.7 & 29.0 & 24.9 \\ \midrule
Ours&\makecell{\textbf{43} \\↓95.0\%}&\makecell{\textbf{0}\\↓100.0\%}&\makecell{\textbf{486}\\↓87.3\%}&\makecell{\underline{384}\\↓81.1\%}&\makecell{\textbf{132}\\↓96.1\%}&\makecell{\underline{201}\\↓70.5\%}&\makecell{\textbf{7}\\↓98.3\%}&\makecell{\textbf{13}\\↓89.4\%}&\makecell{\textbf{1266}\\↓88.8\%} & \textbf{12.1} & \textbf{5.60} & 30.0 & \textbf{21.7} \\ \bottomrule
    \end{tabular}
    \label{tab: nudity detection}
\end{table*}

\begin{table*}[ht]
    \centering
    \renewcommand\arraystretch{1.2}
    \caption{The erasure, generation, and defense performance for erasing the objects church and French horn. R.: The Concept Ratio (\%). S.: The Concept Score. C.: CLIP-Score. F.: FID. P.: P4D. U.: UDAtk.}

    \begin{tabular}{ l | cc | cc |p{0.5cm}<{\centering} p{0.3cm}<{\centering} p{0.3cm}<{\centering} p{0.8cm}<{\centering} || l|  cc | cc | p{0.5cm}<{\centering} p{0.3cm}<{\centering} p{0.3cm}<{\centering} p{0.8cm}<{\centering}
    }
    
    \toprule

     \multirow{2}{*}{Method} & \multicolumn{2}{c|}{Erasure} & \multicolumn{2}{c|}{Generation} & \multicolumn{4}{c||}{Defense} & \multirow{2}{*}{Method} &\multicolumn{2}{c|}{Erasure} & \multicolumn{2}{c|}{Generation} & \multicolumn{4}{c}{Defense} \\
    
     & R.$\downarrow$ & S.$\downarrow$ & C.$\uparrow$ & F.$\downarrow$ & CCE$\downarrow$ & P.$\downarrow$ & U.$\downarrow$ & RAB$\downarrow$
     & & R.$\downarrow$ & S.$\downarrow$ & C.$\uparrow$ & F.$\downarrow$ & CCE$\downarrow$ & P.$\downarrow$ & U.$\downarrow$ & RAB$\downarrow$
     \\

     \midrule
     \multicolumn{9}{c||}{Church} & \multicolumn{9}{c}{French horn}\\ 
     \cmidrule(lr){1-9} \cmidrule(lr){10-18}
     SD \cite{rombach2022high} & 85.8 & 84.5 & 31.5 & 21.1 & 100 & 100 & 100 & 94.1  & SD \cite{rombach2022high} & 99.9 & 99.7 & 31.5 & 21.1 & 100 & 100 & 100 & 98.6 \\ \midrule
    SLD  \cite{schramowski2023safe} & 40.7 & 38.2 & 30.7 & 29.0 & \underline{80.3} & 89.0 & \underline{16.0} & 42.3 & SLD \cite{schramowski2023safe} & \underline{27.3} & 24.8 & 30.2 & 30.9 & 100 & 92.0 & 4.00 &  21.1 \\
     CA \cite{kumari2023ablating} & 69.3 & 67.4 & 31.0 & 26.6 & 91.4 & 94.0 & 80.0 & 21.1 & CA \cite{kumari2023ablating}& 76.0 & 71.1 & \textbf{31.7} & \underline{23.4} & \underline{87.1} & 72.0 & 40.0 & 74.1  \\
     ESD \cite{gandikota2023erasing} & 75.1 & 74.2 & \textbf{31.5} & 25.5 & 95.1 & 93.0 & 91.0 & 93.7 &  ESD \cite{gandikota2023erasing}& 88.9 & 87.7 & 31.5 & 25.5 & 97.2 & 94.0 & 100 & 93.1 \\
     FMN \cite{zhang2023forget} & 81.5 & 79.9 & \underline{31.4} & 28.8 & 100 & 94.0 & 77.0 & 92.3 & FMN \cite{zhang2023forget} & 94.1 & 92.9 & 31.4 & 29.4 & 99.0 & 95.0 & 97.0 & 91.6 \\
     UCE \cite{gandikota2024unified} & \underline{29.1} & \underline{27.7} & 31.3 & 27.0 & 86.4 & 78.0 & 35.0 & 55.5 & UCE \cite{gandikota2024unified} & 37.5 & 35.3 & 31.3 & 26.3 & 97.3 & 83.0 & 17.0 & 36.2 \\
     SalUn \cite{fan2023salun} & 33.8 & 32.7 & 30.9 & \underline{23.2} & 92.0 & \underline{60.0} & 41.0 & \textbf{16.6} & SalUn \cite{fan2023salun} & 28.6 & \underline{17.6} & \underline{31.6} & \textbf{22.1} & 97.0 & \textbf{30.0} & \underline{1.00} & \underline{20.7} \\ \midrule
     Ours & \textbf{26.2} & \textbf{25.1} & 30.6 & \textbf{22.6} & \textbf{29.1} & \textbf{49.0} & \textbf{0.00} & \underline{19.2} & Ours & \textbf{18.0} & \textbf{17.0} & 30.6 & \underline{23.4} & \textbf{36.7} & \underline{36.0} & \textbf{0.00} & \textbf{18.3} \\
    \bottomrule
    \end{tabular}
    \label{tab: erase objects}
\end{table*}

\begin{table*}[!h]
    \centering
    \renewcommand\arraystretch{1.2}
    \caption{The erasure, generation, and defense performance for erasing the painting styles of Van Gogh and crayon. R.: The Concept Ratio (\%). S.: The Concept Score. C.: CLIP-Score. F.: FID. P.: P4D. U.: UDAtk.}

    \begin{tabular}{ l | cc | cc | p{0.5cm}<{\centering} p{0.3cm}<{\centering} p{0.3cm}<{\centering} p{0.8cm}<{\centering} || l|  cc | cc | p{0.5cm}<{\centering} p{0.3cm}<{\centering} p{0.3cm}<{\centering} p{0.8cm}<{\centering}
    }
    
    \toprule

     \multirow{2}{*}{Method} & \multicolumn{2}{c|}{Erasure} & \multicolumn{2}{c|}{Generation} & \multicolumn{4}{c||}{Defense} & \multirow{2}{*}{Method} &\multicolumn{2}{c|}{Erasure} & \multicolumn{2}{c|}{Generation} & \multicolumn{4}{c}{Defense} \\
    
     & R.$\downarrow$ & S.$\downarrow$ & C.$\uparrow$ & F.$\downarrow$ & CCE$\downarrow$ & P.$\downarrow$ & U.$\downarrow$ & RAB$\downarrow$
     & & R.$\downarrow$ & S.$\downarrow$ & C.$\uparrow$ & F.$\downarrow$ & CCE$\downarrow$ & P.$\downarrow$ & U.$\downarrow$ & RAB$\downarrow$
     \\
     \midrule
     \multicolumn{9}{c||}{Van Gogh} & \multicolumn{9}{c}{Crayon}\\ 
     \cmidrule(lr){1-9} \cmidrule(lr){10-18}
     SD \cite{rombach2022high}& 98.5 & 91.7 & 31.5 & 21.1 & 100 & 100 & 100 & 99.9 & SD \cite{rombach2022high} & 95.6 & 71.6 & 31.5 & 21.1 & 100 & 100& 100 & 92.9\\ \midrule
     SLD \cite{schramowski2023safe} & 9.40 & \textbf{4.23} & 30.1 & 30.3 & 30.9 & 17.0 & \textbf{27.0} & 15.3& SLD \cite{schramowski2023safe}& 35.7 & 27.8 & 30.7 & 27.1& 50.7& 80.0& 69.0& 34.3 \\
     CA \cite{kumari2023ablating}& 9.70 & \underline{5.36} & 31.3 & \underline{23.9} & \underline{20.4} & 13.0 & 43.0 & \underline{10.4}& CA \cite{kumari2023ablating} & 23.7& 16.7 &30.8 & 26.8& 5.20 &19.0 &61.0 & 9.40 \\
     ESD \cite{gandikota2023erasing}& 87.2 & 80.8 & \textbf{31.5} & 25.5 & 96.1 & 99.0 & 100 & 99.7& ESD  \cite{gandikota2023erasing} & 89.6 & 66.4 & \textbf{31.5}& \underline{25.4} & 87.4&97.0 &100 & 89.5\\
     FMN \cite{zhang2023forget}& 53.0 & 43.8 & \underline{31.4} & 29.4 & 61.0 & 92.0 & 94.0 & 80.3 & FMN  \cite{zhang2023forget}& 78.0 & 67.0 & 31.3 & 29.7 & 71.0 & 94.0 & 100 & 82.5 \\
     UCE \cite{gandikota2024unified}& 61.8 & 53.5 & \textbf{31.5} & 25.4 & 69.4 & 95.0 & 98.0 & 89.9& UCE  \cite{gandikota2024unified} &54.7 & 39.9 & \underline{31.4} & 25.6& 47.4& 88.0& 100& 53.5\\
     SalUn \cite{fan2023salun}& \underline{8.10} & 7.83 & 30.5 & 26.7 & 24.9 & \underline{12.0} & \underline{32.0} & 13.3 & SalUn \cite{fan2023salun}&\underline{1.40} & \underline{6.47} &19.6 &221 &\underline{4.40} &\textbf{5.00} & \underline{51.0} & \underline{1.00} \\ \midrule
     Ours & \textbf{4.40} & 12.6 & 30.7 & \textbf{22.0} & \textbf{16.0} & \textbf{9.00} & 35.0 & \textbf{8.80} & Ours &\textbf{0.59} & \textbf{6.37}&30.0 & \textbf{24.4}& \textbf{4.00}& \underline{7.00} & \textbf{39.0} & \textbf{0.30}\\

    \bottomrule
    \end{tabular}
    \label{tab: erase styles}
\end{table*}

\begin{figure*}[htbp]
    \centering
    \includegraphics[width=\linewidth]{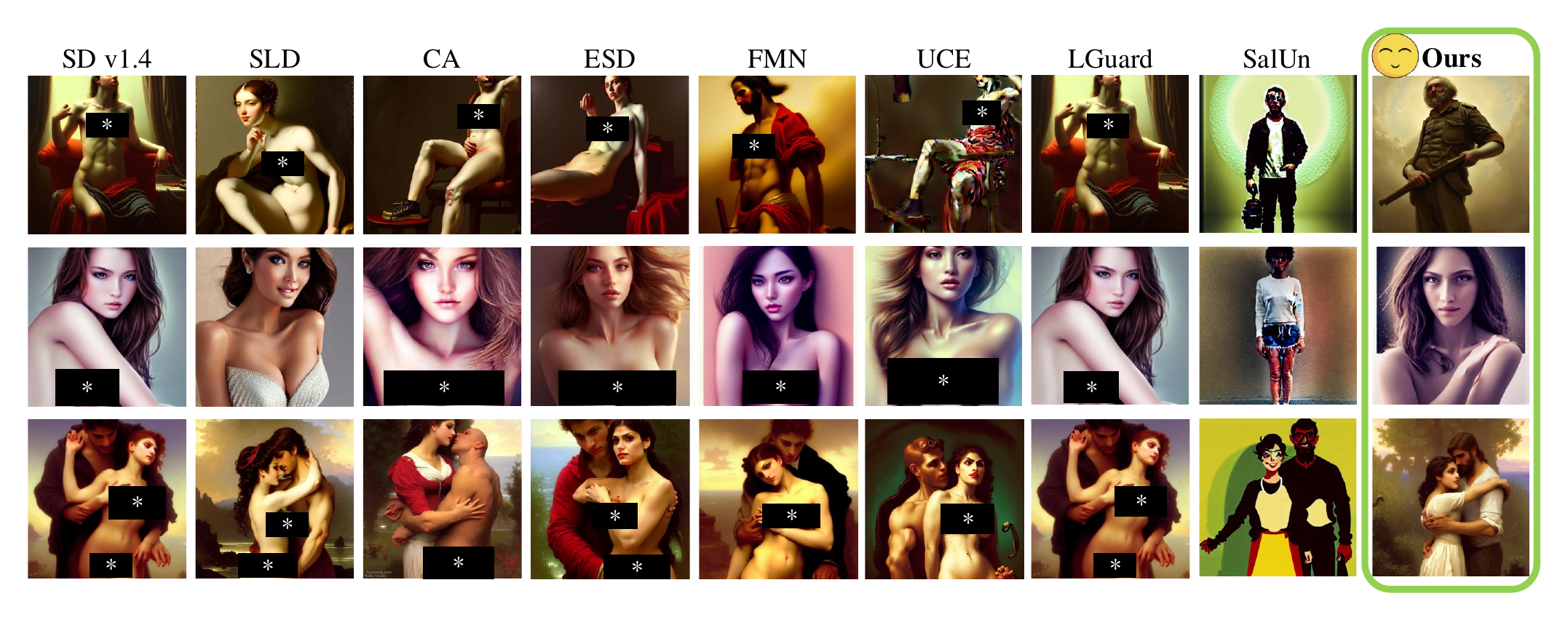}
    \caption{The visualizations of the erasure performance. The images in each row are generated using the same prompts.}
    \label{fig: vis_side_by_side}
\end{figure*}

\begin{table}[t]
    \centering
    \caption{The defense performance for erasing nudity. We report ASR (\%, $\downarrow$) for each attack method.}
    \renewcommand\arraystretch{1.2}
    \begin{tabular}{l|c c c c c}
    \toprule
        \multirow{2}{*}{Method} & \multicolumn{5}{c}{Defense}\\
        \cmidrule(lr){2-6}
        & CCE & P4D & UDAtk & RAB & MMA-Image \\ \midrule
        SD \cite{rombach2022high} &  100 & 100 & 100 & 98.6 & 97.1 \\\midrule
     SLD \cite{schramowski2023safe} & \underline{32.2} & 63.0 & 100 & 94.1 & 24.5 \\
     CA \cite{kumari2023ablating} & 100 & 37.0 & 85.0 & 65.4 & 67.6  \\
     ESD \cite{gandikota2023erasing} & 88.8 & 56.0 & 100 & 98.4 & 96.0 \\
     FMN \cite{zhang2023forget} & 94.3 & 49.0 & 78.0 & 97.3 & 94.6 \\
     UCE \cite{fan2023salun} & 89.6 & 58.0 & 100 & 97.2 & 94.7 \\
     SalUn \cite{fan2023salun} & 47.6 & \underline{36.0} & \underline{37.0} & \underline{28.4} & \underline{19.9} \\
     LGuard* \cite{liu2025latent} & 63.6 & 68.0 & 99.0 & 36.1 & 75.4 \\ \midrule
     Ours & \textbf{27.7} & \textbf{14.0} & \textbf{18.0} & \textbf{26.2} & \textbf{16.0} \\
        \bottomrule
    \end{tabular}
    
    \label{tab: nudity defense}
\end{table}

\subsection{Evaluation Results}
For the erasure and generation performance, we present the results in Tab.\ref{tab: nudity detection} for the nudity concept, Tab.\ref{tab: erase objects} for the object concepts, and Tab.\ref{tab: erase styles} for the style concepts. Compared with other methods, Dark Miner achieves better results on the Concept Ratio for erasing all the concepts. It demonstrates that our method has the best performance in preventing generation conditioned on the user prompts, regardless of whether the concept is nudity, an object, or a style. For the Concept Score, Dark Miner also obtains the best performance in most cases. It indicates that the images generated by the model erased by our method exhibit a higher degree of divergence from the intended concepts in comparison to other methods.

For the concept of nudity, Tab.\ref{tab: nudity detection} presents the detailed detection results of the classes. The results show that our method produces the most obvious erasing effect on each class, except for the belly and male breast classes. These two classes have the lowest level of nudity among sexual content, especially compared to the genitalia classes. Therefore, compared with previous methods, Dark Miner is the most effective method for erasing key nudity elements. We also observe an unbalanced performance on the nudity classes for various erasure methods, including ours. This discrepancy can likely be attributed to biases present both in our image pool and the I2P evaluation set. To address this shortcoming, we can create an image pool containing the specific nudity class for which we want to enhance the erasure performance. It can enable the more effective erasure of the corresponding nudity class. We provide the side-by-side comparison visualizations in Fig.\ref{fig: vis_side_by_side}.

For the defense performance, we present the results in Tab.\ref{tab: nudity defense} for the nudity concept, Tab.\ref{tab: erase objects} for the object concepts, and Tab.\ref{tab: erase styles} for the style concepts. The related results show that most of the existing methods fail to guarantee the erasure of concepts when prompting the models with adversarial prompts. For example, when erasing nudity, CA reduces the Concept Ratio by 32\% but achieves an ASR of 100\% under the attack CCE. When erasing church, SLD reduces the Ratio by 45.1\% but achieves an ASR of 89.0\% under the attack P4D. On the contrary, our method achieves the lowest ASR on most attacks when erasing these concepts. It demonstrates that our method has better defense ability against attacks.

Besides the individual implementation of the above attacks, we further consider a more challenging attack, AutoAttack (also named Attack Ensemble) \cite{croce2020reliable}. It refers to the attack that runs its component attacks and chooses the best. We conduct the experiments on the erasure of nudity using the prompt set collected by the official implementation of RAB \cite{tsai2023ring}. For CCE \cite{pham2023circumventing}, the images for concept inversion are the ones generated by the corresponding prompts using the original diffusion models. The results show that the ASR of our method against this attack is 28.0\%, which is lower than CA (76.0\%), SalUn (34.0\%), and LatentGuard (42.0\%). It further provides evidence that our method has better robustness.

\begin{figure*}[t]
  \centering
  \includegraphics[width=\linewidth]{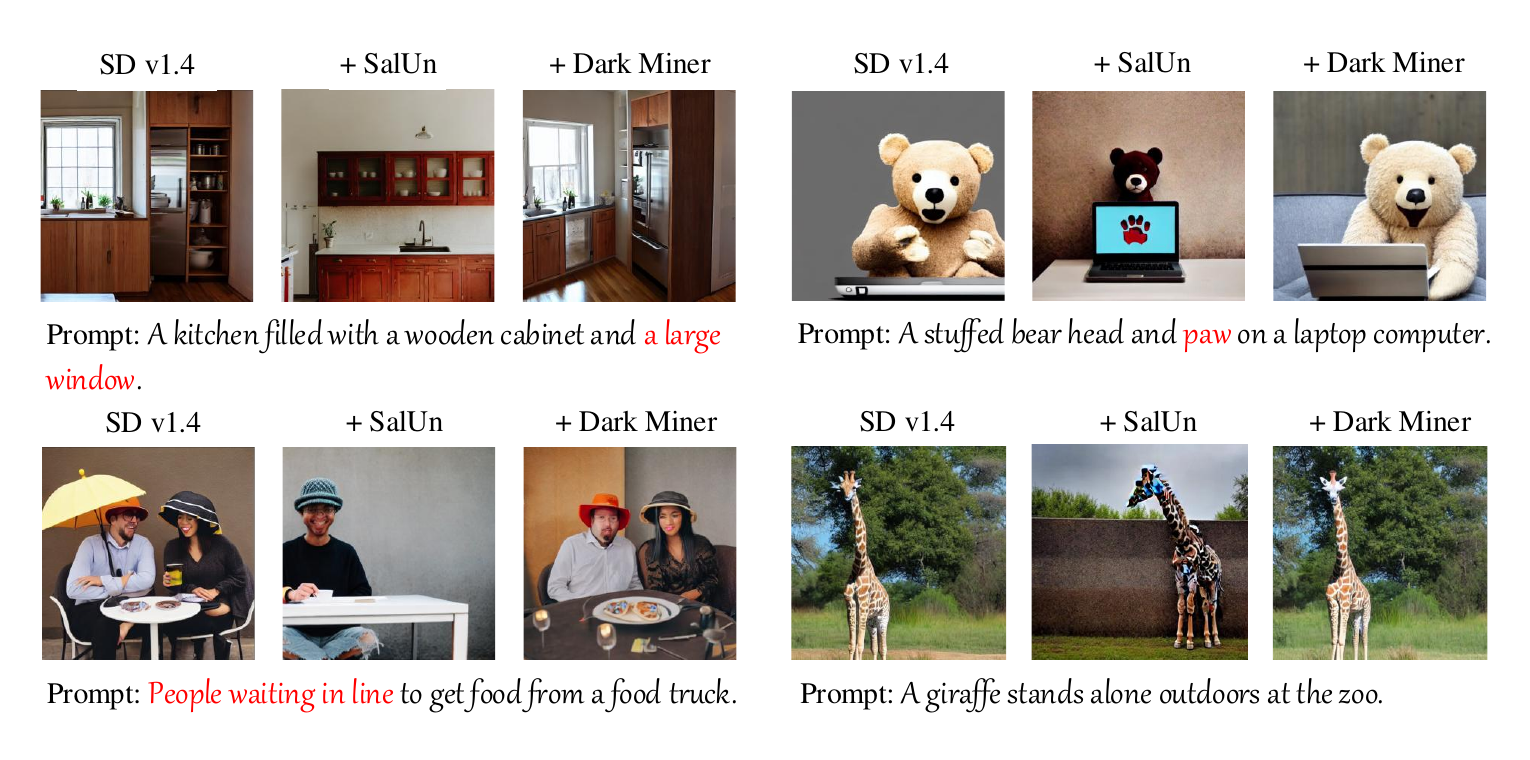}
  \caption{The visualizations of the generation performance. The erased concept is nudity. The red font denotes the missed words caused by SalUn.}
  \label{fig: tsamples}
\end{figure*}

These evaluation results also reveal that SalUn is second only to our method in multiple metrics for the erasure and defense performance. However, it often sacrifices the generation performance for better erasure and defense performance. When erasing nudity and the painting styles, its generation performance is the worst, with CLIP and FID lower than the original model and other methods significantly. To illustrate it intuitively, in Fig.\ref{fig: tsamples}, we show some images generated by the models with nudity erased by SalUn and ours. It can be seen that the images generated by SalUn have obvious inconsistencies with the corresponding prompts. For example, for the prompt ``\textit{A kitchen filled with a wooden cabinet and a large window}'', the image generated by SalUn misses ``\textit{the large window}''. On the contrary, our proposed method can still accurately generate images consistent with the given texts. In addition, SalUn leads to a performance drop in image quality. In the last example of Fig.\ref{fig: tsamples}, the pattern of the giraffe is obviously distorted.

\subsection{Ablation Studies}

\begin{figure}
    \centering
    \includegraphics[width=0.7\linewidth]{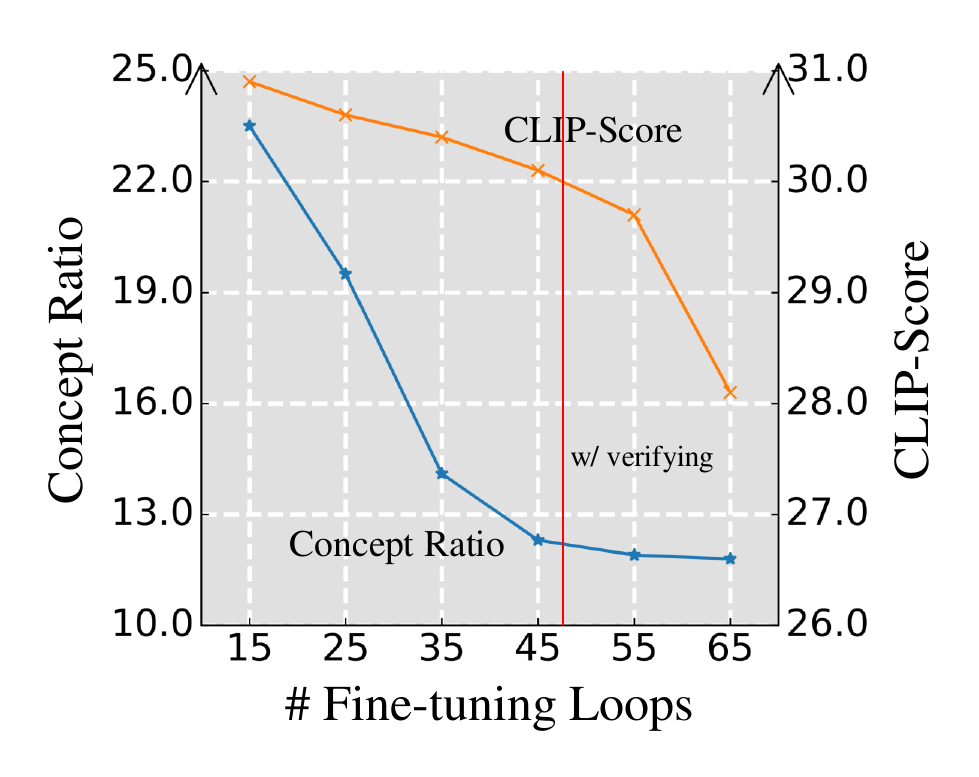}
    \caption{The erasure performance (Concept Ratio, \%↓) and the generation performance (CLIP-Score, ↑) when we ablate the verifying step and set the fine-tuning loops manually. The erased concept is nudity. The red line indicates the ending point determined by the verifying step.}
    \label{fig:ablate_verifying}
\end{figure}

\textbf{Ablation of Dark Miner Steps.} First, the mining step is ablated. We experiment on the concept of nudity and set the input prompt for the fine-tuning as “nudity”, aligning with the setting in \cite{gandikota2023erasing}. With other settings fixed, the results show that the concept ratio is 41.9\%. As a comparison, our method with the mining step achieves 12.1\%. It demonstrates that the mining step erases concepts in the model more thoroughly. Second, the verification step is ablated, and we set the fine-tuning loops manually. Figure \ref{fig:ablate_verifying} shows how the performance changes with the number of fine-tuning loops. Setting a small number of training loops results in a higher concept ratio, i.e., incomplete erasure. On the contrary, setting a large number of training loops results in the model being unable to stop fine-tuning in time when the mining step fails, resulting in excessive damage to the model generation, but almost unchanged nudity ratio. Therefore, it is necessary to adaptively control the model fine-tuning process with the proposed verifying step. It should be noted that we cannot ablate the circumventing step because we cannot update the model and erase concepts without the fine-tuning in this step.

\begin{table}
\centering
    \normalsize
    \captionof{table}{The ablation results in embedding lengths. The erased concept is nudity.}
    \begin{tabular}{c | c c}
    \toprule
     Embedding Length & Ratio $\downarrow$ & CLIP $\uparrow$  \\
     \midrule
     1 &  19.65 & 30.97 \\
     8 &  15.85 & 30.55 \\
     16 &  12.13 & 30.11\\
     32 & 12.07 & 30.00 \\
    \bottomrule
    \end{tabular}
    \label{tab: length}
\end{table}

\begin{table}[t]
 \centering
    \normalsize
    \captionof{table}{The ablation results in anchor prompts. The erased concept is nudity.}
    \label{tab: prompt}
    \begin{tabular}{c | c c}
    \toprule
     Anchor Prompt & Ratio $\downarrow$ & CLIP $\uparrow$  \\
     \midrule
     \small $\varnothing$ &  12.23 & 30.62 \\
     \small a natural photo &  12.13 & 30.12 \\
     \small a happy photo &  12.36 & 30.87\\ 
     \small a photo & 12.07 & 30.00 \\
    \bottomrule
    \end{tabular}
\end{table}

\textbf{Embedding lengths.} Tab.\ref{tab: length} shows the effect of the embedding lengths on the erasure performance. A short embedding cannot capture enough semantic representations of concepts in the mining stage. It leads to the early stopping of Dark Miner and therefore incomplete erasure.

\textbf{Anchor prompts.} The results using different anchor prompts ($c_0$ in Eq.\ref{eq: circumvent}) are shown in Tab.\ref{tab: prompt}. Overall, their results are similar. The results of the empty prompt and ``\textit{a happy photo}'' are slightly inferior to others. We speculate that the reason for the empty prompt may be that the generated images are more random, leading to divergence in the optimization direction. The reason for ``\textit{a happy photo}'' may be that the generated images usually contain people and people are often associated with the concept of nudity, leading to an incomplete erasure. This point inspires us that selecting anchor prompts should ideally be tailored to the specific target concept.

\begin{table}[t]  
\normalsize
    \centering
    \caption{The results with different image pools (erase nudity). The images in the pools are generated with different random seeds. 2024 is used in the paper. } 
    \normalsize
    \begin{tabular}{c|c c c c}
    \toprule
       \multirow{2}{*}{Seed} & Erasure & Generation & \multicolumn{2}{c}{Defense} \\
       & Ratio (\%) & CLIP & RAB (\%)& CCE (\%)\\ \midrule
       2024  & 12.07 & 30.00 & 26.21 & 27.67 \\
       2020 & 12.04 & 29.98 & 24.39 & 27.39 \\
       2028 & 12.38 & 30.34 & 26.79 & 28.09\\ \midrule
       Avg. & 12.16  & 30.11 & 25.80 & 27.72 \\
       Std. & 0.15 & 0.17 & 1.02 & 0.29 \\
    \bottomrule
    \end{tabular}
    \label{tab: pool}
\end{table}

\begin{table}[t]
\normalsize
\centering
    \captionof{table}{The results with different image pool sizes. The erased concept is nudity.}
    \normalsize
    \begin{tabular}{c|c  c}
    \toprule
       Image Pool Size  & Ratio $\downarrow$ & CLIP $\uparrow$   \\ \midrule
       20  & 21.45 & 30.79\\
       200 & 12.07 & 30.00 \\
       2000 & 11.84 & 29.95 \\
    \bottomrule
    \end{tabular}
    \label{tab: pool size}
\end{table}

\begin{table}[t]
\normalsize
 \centering
    \captionof{table}{The ablation results when ablating the preservation terms in Eq.\ref{eq: preservation}. The erased concept is nudity.}
    \begin{tabular}{c c c | c c }
    \toprule
        \multicolumn{3}{c|}{Ablation Term} & \multirow{2}{*}{Ratio$\downarrow$} & \multirow{2}{*}{CLIP$\uparrow$}  \\ 
        \cmidrule(lr){1-3}
        $c_0$ & $0c$ & $-c$ & & \\
        \midrule
       \ding{55} & \ding{52} & \ding{52} & 10.00 & 29.31 \\
       \ding{52} & \ding{55} & \ding{52} & 12.51 & 29.92 \\ 
       \ding{52} & \ding{52} & \ding{55} & 8.24 & 28.73 \\ \midrule
       \ding{52} & \ding{52} & \ding{52} & 12.07 & 30.00 \\
    \bottomrule
    \end{tabular}
    \label{tab: ablation preservation}
\end{table}

\textbf{Image pools.} An image pool is required by our method to optimize embeddings. We conduct two analyses on image pools. First, we use different random seeds to generate images for the image pool while maintaining the sampling sequence and other settings. The results are shown in Tab.\ref{tab: pool}. Their small standard deviations on the metrics indicate that our method is robust to different image pools. Next, we selectively sample 20, 200, and 2000 generated images to build the image pool, with the results presented in Tab.\ref{tab: pool size}. Other configurations are fixed. On the one hand, a smaller size constrains the mining capability, leading to a sampling bias where certain crucial elements of nudity, such as anus and male genitalia, may be omitted. Despite this limitation, the erasure performance using just 20 images still surpasses most methods, as evident from the comparisons in Table \ref{tab: nudity detection}. On the other hand, as the pool size grows larger, the performance improvement diminishes. Specifically, when the size increases from 200 to 2000 (a 10x increase), the performance only increases by 0.3 (about a 0.025x increase). This is attributed to the fact that the training stops before many images in the pool are sampled. Considering these factors, dynamically sampling images and expanding the image pool during training is an effective strategy.

\textbf{Embedding Pools.} To demonstrate the role of the embedding pools introduced in our method, we conduct this ablation experiment on the concept of nudity. Specifically, in each training loop, only the embeddings mined in the current loop are used to fine-tune the model. The results show that the Ratio increases slightly by about 1.5\%, and the training time is significantly prolonged. The number of training loops is extended from 48 to 65. It suggests that the over-fitting to the current embedding in each training loop compromises the unlearning effect of previously mined embeddings, which ultimately leads to a decrease in training efficiency.

\textbf{Preservation terms.} In Eq.\ref{eq: preservation}, three terms are used to preserve the generative ability of the model. We ablate these terms respectively and discuss their effectiveness. Other configurations are fixed. The results are shown in Tab.\ref{tab: ablation preservation}. The results show that $-c$ is the most important preservation term. During the fine-tuning, changes in $c$ results in corresponding changes in $-c$. As a result, the term $-c$ helps protect more irrelevant embeddings. $c_0$ and $0c$ can help improve the generation performance but the effect is weaker than $-c$. Furthermore, our observation reveals that eliminating certain preservation terms results in enhanced erasure performance. This is attributed to the absence of preservation constraints allowing for more erasures at the cost of some generative capabilities.

\textbf{Verifying thresholds.} We conduct the experiments using different verifying thresholds and the results are shown in Tab.\ref{tab: thr}. The erased concept is nudity. It can be seen that the erasure performance increases as the threshold decreases. Overall, a high threshold will cause Dark Miner to stop early, resulting in an incomplete erasure.

\textbf{Verifying Tools.} In Dark Miner, we use CLIP as the verifying tool to identify images with the target concepts, without training a specific model to recognize each target concept. To demonstrate that it does not compromise the erasure performance, we replace CLIP with NudeNet in the verifying step to erase nudity in text-to-image diffusion models. The results demonstrate that they both stop the fine-tuning process at the same point, leading to the same results. More discussions will be conducted in Sec.\ref{sec: Verifying Using CLIP}.

\begin{table}[t]
    \centering
    \caption{The ablation results when using different verifying thresholds $\tau$ for Eq.\ref{eq: verify score} (erase nudity). We report the number of training loops (\# loops), the training time (hour), the erasure performance (Ratio, \%), the generation performance (CLIP), and the defense performance (ASR, \%) under the attack RAB.}
    \normalsize
    \begin{tabular}{c| cc | c c c c}
    \toprule
       Threshold & \# loops & Time & Ratio$\downarrow$ & CLIP $\uparrow$ & RAB$\downarrow$ \\ \midrule
       0.4  & 15 & 18.5 & 23.5 & 30.9 & 29.0 \\
       0.3 &  20 & 24.8 & 21.0 & 30.8 & 26.3  \\
       0.2 &  48  & 59.3 & 12.1 & 30.0 & 26.2  \\ 
    \bottomrule
    \end{tabular}
    \label{tab: thr}
\end{table}

\subsection{Discussions}
\subsubsection{Rethinking Dark Miner}
In Sec.\ref{sec: method}, we provide a mathematical perspective on Dark Miner, namely, minimizing a tight upper bound on the probability of concept generation. It should be noted that Dark Miner can also be interpreted as a form of adversarial training. This is because of the alternative execution of the mining and circumventing steps. The mining step continuously searches for embeddings associated with target concepts that have yet to be erased.

\subsubsection{Verifying Using CLIP}
\label{sec: Verifying Using CLIP}
In Dark Miner, we design a verifying step to determine whether to continue the training process using the CLIP delta feature. The previous work \cite{lyu2023deltaedit} uses it in the task of image manipulation. In this scenario, the input images for feature calculations are usually paired, e.g., an image of a girl with yellow hair and an image of a girl with black hair. Compared with this scenario, the image elements in concept erasure are more diverse and do not appear in pairs, thereby requiring a new study to confirm the availability. In this section, we demonstrate its effect in identifying concepts in images. Using SD v1.4, we sample 100 images using the prompt ``\emph{a photo}'', and 100 images using the prompt ``\emph{a photo of [CONCEPT]}''. For each concept, we use each of the former images as the reference image, and each of the latter as the target image. Then these images with/without the concept are regarded as the "positive" and "negative" classes, respectively. We calculate the proposed metrics for these images. In total, there are 2*100*100*100=2,000,000 pairs of samples. We use these sample pairs to calculate the Area Under Curve (AUC). The results are shown in Tab.\ref{tab: clip auc}. The results demonstrate that our method can help identify images effectively.

\begin{table}[t]
    \centering
    \caption{The identification performance of concepts using our proposed verifying method. AUC: the Area Under Curves.}
    \normalsize
    \begin{tabular}{l|c}\toprule
        Concept & AUC \\ \midrule
        Nudity & 0.990 \\
        Church & 0.989\\
        French Horn   & 1.000\\
        Van Gogh's painting style & 0.997\\
        Crayon painting style   & 0.960\\ \bottomrule
    \end{tabular}
    \label{tab: clip auc}
\end{table}

\begin{figure}[t]
    \centering
    \includegraphics[width=1\linewidth]{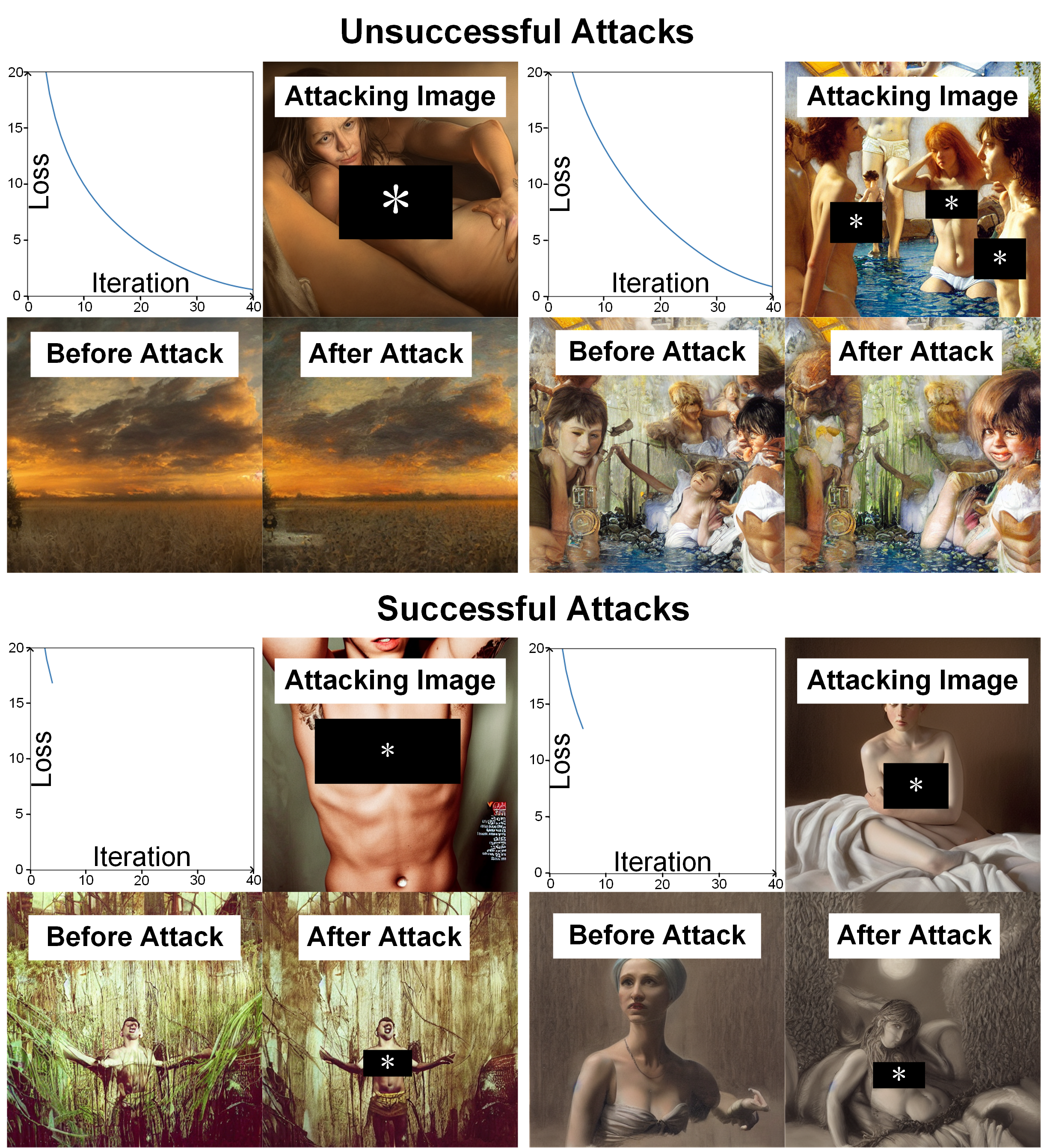}
    \caption{The examples of the successful and unsuccessful attacks by UDAtk. The attacking images are images generated by the original model and used in the attacking process.}
    \label{fig: UD loss}
\end{figure}

\subsubsection{Attack Analysis}
\label{sec: attack analysis}

For attacks like UDAtk, they optimize prompts by gradient back-propagation. In this section, we analyze the optimization process of UDAtk. Specifically, we randomly select two successful and unsuccessful attacks. Fig.\ref{fig: UD loss} shows their loss curves and the images before/after attacking. The original images used for attacking are also shown.

Ideally, the loss curves for successful attacks should decrease, while the loss curves for unsuccessful attacks should be non-decreasing. In Fig.\ref{fig: UD loss}, we can see that for the successful attacks, the loss continues to drop. However, for the unsuccessful attacks, interestingly, the loss also shows a decreasing trend.

We analyze the possible reason for this phenomenon. We find that the generated images and the attacking images are significantly different. Recall the principle of UDAtk. It optimizes prompts by minimizing MSE between real and predicted noises for the noised attacking images. The previous study \cite{balaji2023ediff} reveals that the generation strongly relies on input images during later sampling. Unfortunately, the original images used for attacking are not seen in the evaluation phase. Without their guidance, adversarial prompts successful in training fail in evaluation. Despite success in evaluation, the inappropriateness degree is much lower than in attacking images. We hope that this preliminary analysis will facilitate research on attack methods.

\begin{table}[t]
    \centering
    \caption{The results with more Stable Diffusion models (erase nudity).}
    \normalsize
    \begin{tabular}{l | c c c c}
    \toprule
        Model & Ratio($\downarrow$) & CLIP($\uparrow$) & RAB($\downarrow$) & CCE($\downarrow$) \\ \midrule

       SD v2.0 & 35.3 & 31.7 & 94.2 & 100.0 \\
       + Dark Miner & 8.20 & 30.2 & 20.1 & 24.6\\ \midrule
       SD XL v1.0 & 24.7 & 31.8 & 48.9 & 100.0 \\
       + Dark Miner & 6.9 & 31.2 & 19.7 & 16.3 \\ \midrule
       PixArt-$\alpha$-512 & 1.2 & 31.5 & 0.0 & 1.1 \\
       + Dark Miner & 1.2 & 31.5 & 0.0 & 1.1\\ 
    \bottomrule
    \end{tabular}
    \label{tab: sd version}
\end{table}

\subsubsection{Application Scope of Dark Miner}
Dark Miner is designed specifically for white-box scenarios and has no special restrictions on the model architectures, thus exhibiting a wide range of model applicability. The mining step optimizes input embeddings, imposing only the requirement that the models are differentiable. The verifying step operates on generated images and is independent of the diffusion models. The circumventing step involves inserting and fine-tuning LoRA modules in the attention layers. Fortunately, the attention mechanism is a foundational component across text-to-image diffusion architectures, and LoRA tuning is a standard lightweight adaptation technique. Consequently, the method theoretically exhibits a broad model applicability, provided that access to the model parameters and gradients. With other settings fixed, we erase the nudity for SD v2.0, SD XL v1.0, and PixArt-$\alpha$-512, as shown in Tab.\ref{tab: sd version}. In addition to its effective performance on SD, an interesting observation is that when it is applied to PixArt-$\alpha$-512, a model that inherently has a low nudity ratio, it terminates after the first verifying step. This suggests that preparing a cleaner model can help reduce time costs.

\subsubsection{Potential Society Impacts}
\label{sec: society}

This work will have a beneficial impact on our society. In the era of AIGC, multiple open-source and commercial generative models are readily accessible to users, allowing individuals to effortlessly obtain generated images. Nonetheless, given the extensive training datasets, generative models may inevitably produce undesirable images, such as nudity and protected copyrights. Furthermore, malicious users can employ attack methodologies to prompt models into generating inappropriate content. To tackle this challenge, we carry out this study to defend against undesirable generations, encompassing those triggered by diverse attack methods. We aspire that this work will contribute to fostering a safer generation in the AIGC landscape.
\section{Conclusion}

For erasing concepts in text-to-image diffusion models, most methods focus on modifying the generation distributions conditioned on collected related texts. However, they often cannot guarantee the desired generation of prompts unseen in the training phase, especially the adversarial prompts. In this paper, we analyze this task and point out that they fail to minimize the probabilities of undesirable generation from a global perspective, leading to an overall likelihood that is not sufficiently weakened. To address this challenge,  we propose Dark Miner. It mines embeddings with the maximum generation likelihood of the target concepts and circumvents them, reducing the total probability of generation. Experiments show that compared with the previous methods, our method exhibits the best erasure and defense performance in most cases while preserving the generation capability.

\bibliographystyle{IEEEtran}
\bibliography{IEEEabrv, ref}



\end{document}